\documentclass[letterpaper]{article} % DO NOT CHANGE THIS
\usepackage{aaai25}  % DO NOT CHANGE THIS
\usepackage{times}  % DO NOT CHANGE THIS
\usepackage{helvet}  % DO NOT CHANGE THIS
\usepackage{courier}  % DO NOT CHANGE THIS
\usepackage[hyphens]{url}  % DO NOT CHANGE THIS
\usepackage{graphicx} % DO NOT CHANGE THIS
\usepackage{natbib}  % DO NOT CHANGE THIS AND DO NOT ADD ANY OPTIONS TO IT
\usepackage{caption} % DO NOT CHANGE THIS AND DO NOT ADD ANY OPTIONS TO IT
\usepackage{algorithm}
\usepackage{algorithmic}

\usepackage{newfloat}
\usepackage{listings}

\usepackage{algorithm}
\usepackage{algorithmic}
\usepackage{amssymb}% http://ctan.org/pkg/amssymb
\usepackage{pifont}% http://ctan.org/pkg/pifont
\newcommand{\cmark}{\ding{51}}%
\newcommand{\xmark}{\ding{55}}%

\usepackage{booktabs}

\newcommand{\mpage}[2]
{
\begin{minipage}{#1\linewidth}\centering
#2
\end{minipage}
}

\usepackage{multirow}

\DeclareCaptionStyle{ruled}{labelfont=normalfont,labelsep=colon,strut=off} % DO NOT CHANGE THIS
\floatstyle{ruled}
\newfloat{listing}{tb}{lst}{}
\floatname{listing}{Listing}
\title{HiCM$^2$: Hierarchical Compact Memory Modeling for Dense Video Captioning}
\author{
    Minkuk Kim\textsuperscript{\rm 1}, Hyeon Bae Kim\textsuperscript{\rm 1}, Jinyoung Moon\textsuperscript{\rm 2}, Jinwoo Choi\textsuperscript{\rm 1}\equalcontrib, Seong Tae Kim\textsuperscript{\rm 1}\equalcontrib\\
}
\affiliations{
    \textsuperscript{\rm 1}Kyung Hee University, Republic of Korea\\
    \textsuperscript{\rm 2}Electronics and Telecommunications Research Institute (ETRI), Republic of Korea\\
    \{asdjklfgh97,hyeonbae.kim\}@khu.ac.kr, jymoon@etri.re.kr, \{jinwoochoi,st.kim\}@khu.ac.kr

}
\usepackage{bibentry}
\begin{document}

\maketitle
% \footnotetext[1]{Corresponding Authors}
\begin{abstract}
With the growing demand for solutions to real-world video challenges, interest in dense video captioning (DVC) has been on the rise. DVC involves the automatic captioning and localization of untrimmed videos. Several studies highlight the challenges of DVC and introduce improved methods utilizing prior knowledge, such as pre-training and external memory. In this research, we propose a model that leverages the prior knowledge of human-oriented hierarchical compact memory inspired by human memory hierarchy and cognition. To mimic human-like memory recall, we construct a hierarchical memory and a hierarchical memory reading module. We build an efficient hierarchical compact memory by employing clustering of memory events and summarization using large language models. Comparative experiments demonstrate that this hierarchical memory recall process improves the performance of DVC by achieving state-of-the-art performance on YouCook2 and ViTT datasets.
\end{abstract}


\section{Introduction}
\label{sec:Intro}

Dense video captioning (DVC) is a task that involves temporally localizing event boundaries in untrimmed videos and describing each event in natural language. Compared to video captioning\cite{lin2022swinbert,luo2020univl,seo2022end,7984828,pei2019memory,qi2019sports,wang2018reconstruction,chen2017video,pan2016jointly,rohrbach2013translating,venugopalan2014translating,venugopalan2015sequence,tu2021enhancing,tu20222,tu2023relation}, which generates a single caption for a video clip, DVC presents considerable challenges. This task requires not only generating captions but also automatically localizing multiple important event segments within untrimmed videos. In other words, DVC should accurately identify event boundaries and describe each event segment in natural language, making it a far more complex process.

\begin{figure}[t]
    \centering
    \includegraphics[width=0.99\linewidth]{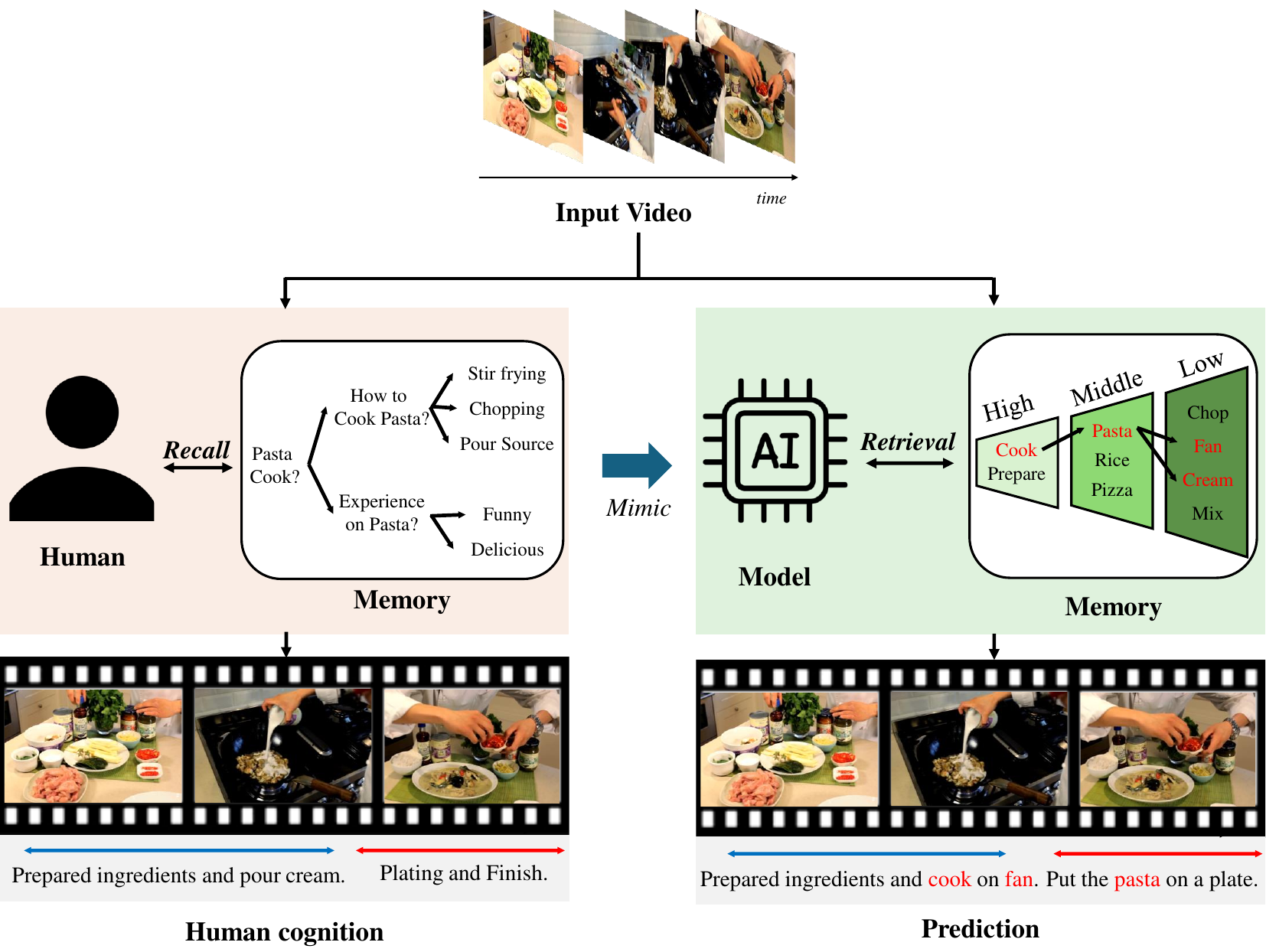}
    \caption{Conceptual figure of the proposed hierarchical compact memory construction. Our method leverages relevant clues from hierarchical compact memory with cross-modal retrieval, effectively mimicking human memory processes for improved event localization and description.}
    \label{fig:concept}
\end{figure}
Recent advancements in vision and language learning have demonstrated remarkable outcomes in cross-modal correlation tasks \cite{radford2021learning,saharia2022photorealistic,li2022blip}. Nevertheless, linking natural language with video remains challenging, largely due to the complexities involved in capturing spatiotemporal information \cite{hayes2022mugen}. Video and language learning demands intricate model architectures, tailored training protocols, and significant computational resources~\cite{cheng2023vindlu}. DVC involves the additional challenge of associating untrimmed videos with natural language to both localize and describe events~\cite{krishna2017dense,wang2021end,yang2023vid2seq}. To address these challenges, several studies have been explored to leverage prior knowledge through a pre-training and fine-tuning approach \cite{yang2023vid2seq,wu2024dibs}.
In addition, a study has been proposed \cite{kim2024you} that enhances DVC by utilizing information stored in memory as an additional source of prior knowledge.
This study is inspired by the observation that humans store and recall information hierarchically. Humans have the ability to identify significant events and describe them by retrieving relevant memories based on cues they have encountered. In cognitive information processing, this phenomenon is called cued recall~\cite{allan1997event,rugg1998neural}. Notably, when recalling, humans tend to first perceive abstract semantic information, then associate it with related categories in memory, and ultimately retrieve specific episodes connected to the recognized information, forming a memory hierarchy~\cite{tulving1972episodic}. By recalling memories and drawing on prior knowledge, humans can describe a given situation based on a rich pool of experiences.

Building on this insight, we introduce a novel DVC model, HiCM$^2$ (Hierarchical Compact Memory Modeling for DVC). Our model effectively recalls relevant episodes and abstract concepts from a hierarchical and compact external memory, enhancing the quality of captions, as in Figure \ref{fig:concept}. To mimic the human recalling process, we design an external memory based on knowledge information extracted from the training data. This knowledge information is organized into a hierarchical memory through iterative clustering using a bottom-up approach. During this process, knowledge information at each level is summarized using large language models (LLMs), transforming it into consolidated memory knowledge. This approach combines hierarchical memory and efficient compact memory construction, resembling human memory organization.

Subsequently, the proposed model extracts potential event candidates from untrimmed videos and retrieves relevant information from the external memory to provide diverse and semantically rich data. Unlike the bottom-up memory construction, information retrieval follows a top-down approach, starting from high-level abstract information and moving towards lower-level details. We first select relevant information from the higher levels and then iteratively refine the search by focusing on related memories from the lower levels, thus facilitating an efficient hierarchical memory read. 

Our main contributions can be summarized as:
\begin{itemize}
    \item  Inspired by the human cognitive process, we introduce a novel DVC method that utilizes cross-modal retrieval from the hierarchical compact memory bank. This approach combines hierarchical organization with compact representations, enabling effective recall of relevant episodes and abstract concepts while minimizing redundancy. To the best of our knowledge, this is the first study to design hierarchical memory structures for DVC.
    
    \item To efficiently extract and utilize hierarchical information, we propose a novel hierarchical memory retrieval method. Our approach leverages a top-down retrieval strategy, starting from high-level abstract information and progressively accessing detailed lower-level data. This method ensures that meaningful and relevant information is effectively retrieved from different levels of the hierarchical memory.
    
    \item To demonstrate the effectiveness of our hierarchical memory retrieval method, we have conducted comprehensive experiments on YouCook2 and ViTT datasets. Our model achieves state-of-the-art on two DVC datasets. Experimental results show that our method could significantly improve the performance by exploiting relevant information from hierarchical memory.
\end{itemize}
%%%%%%%%%%%%%%%%%%%%%%%
\section{Related Work}
\label{sec:RelatedWork}

\subsection{Dense Video Captioning}
% DVC 전반
DVC tasks comprise localizing events within untrimmed videos and generating descriptive captions for each localized event \cite{krishna2017dense}.
Traditionally, a two-stage approach has been employed, where events in the video are first localized, and then captions are generated for each event \cite{iashin2020better, iashin2020multi}. This "localize-then-describe" strategy addresses the two sub-tasks independently. 
However, this two-stage approach has a limitation in that it fails to consider the interaction between the event localization and event captioning tasks, as they are performed independently. To address this issue, recent research has introduced end-to-end approaches that enable joint learning by considering the interaction between the two tasks. 
PDVC \cite{wang2021end} defines DVC as a set prediction task that predicts event intervals and captions simultaneously, using the DETR \cite{carion2020end} architecture to jointly perform event localization and captioning in parallel. 
Vid2Seq \cite{yang2023vid2seq}, on the other hand, defines DVC as a sequence-to-sequence problem, adopting an end-to-end approach that outputs a single sequence containing both timestamps and captions. 
This research also incorporates speech-to-text technology, using transcribed audio information as input to leverage multimodal data. \cite{zhou2024streaming} proposes a visual memory and streaming decoding approach. DIBS \cite{wu2024dibs} introduces a new pretraining framework.
CM$^2$ \cite{kim2024you} applies retrieval-augmented generation (RAG), inspired by how humans process visual information, retrieving relevant information from an external memory pre-constructed with visual data to provide additional context to the model. However, when constructing the external memory, all captions from the target benchmark dataset are included, leading to the potential inclusion of redundant and noisy information.
To address this, our research aims to construct an external memory that filters out redundant or irrelevant information by utilizing clustering and LLMs.

\begin{figure*}[t]
    \centering
    \includegraphics[width=0.77\linewidth]{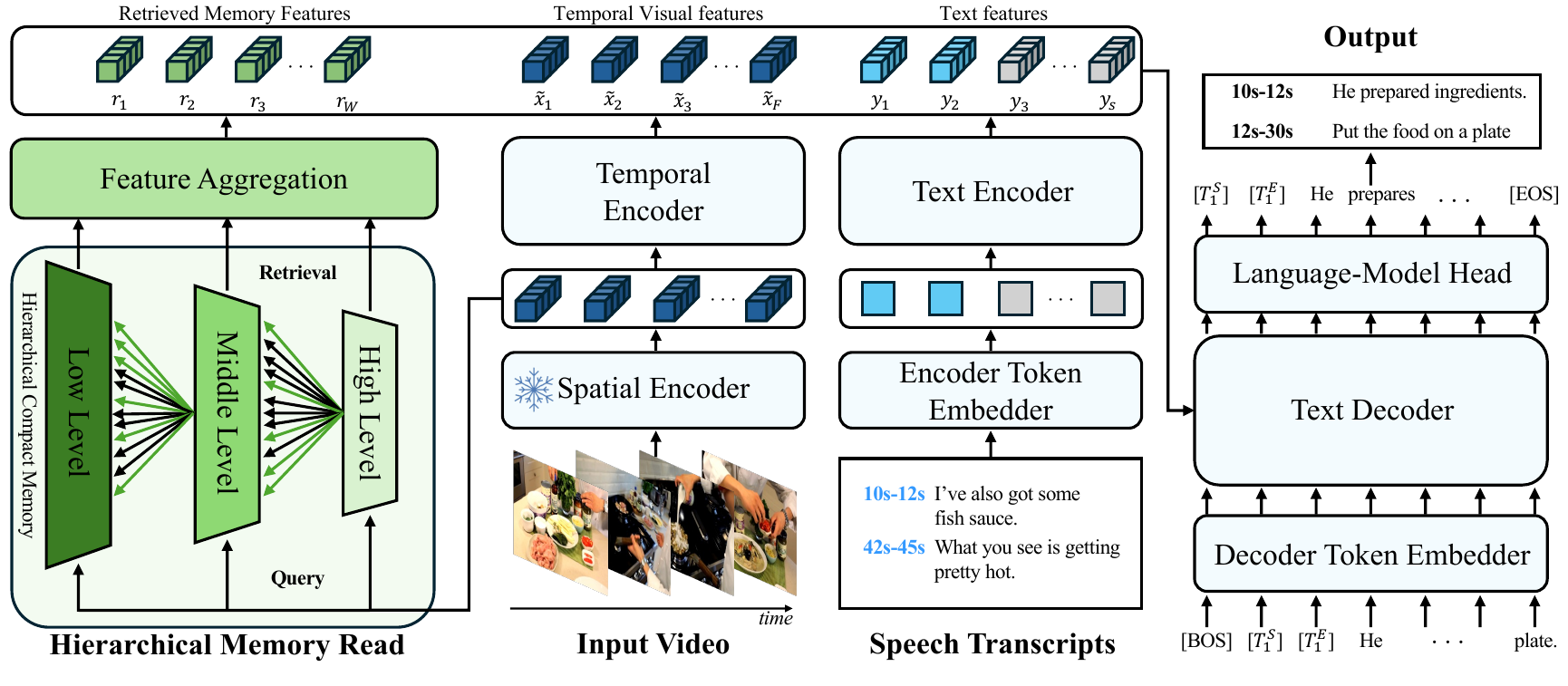}
    \caption{\textbf{Overview of HiCM$^2$.} We show the overall architecture. HiCM$^2$ approaches the DVC task in a memory retrieval-augmented generation manner using hierarchical compact memory. We conduct video-to-text cross-modal retrieval at each hierarchical level using input video features obtained through a pre-trained spatial encoder. The visual features are temporally encoded through a temporal encoder. The time information of speech is converted into time tokens for time tokenization, and the speech text is tokenized for speech encoding. Each encoded feature vector is concatenated and fed into the cross-attention layer of the text decoder. Finally, we obtain a sequence consisting of the start time, end time, and caption.}
    \label{fig:overall}
\end{figure*}

\subsection{Retrieval-Augmented Generation}

Recently, RAG has gained significant popularity for its effective application in LLMs, where it references external knowledge to address issues such as hallucinations \cite{lewis2020retrieval}. 
RAG is particularly effective at reducing the generation of factually incorrect content, making it a valuable approach in various visual captioning tasks \cite{gao2023retrieval}. 
This approach involves retrieving relevant information from an external memory bank based on visual data and providing it to the model \cite{ramos2023smallcap, kim2024you}. 
However, the performance of RAG depends on how constructed the memory and it has not been explored to efficiently construct memory banks for effective DVC. 
In this study, we designed a system to effectively retrieve semantic information, positively impacting model performance. 
To achieve this, we employed a clustering and compact representation method to reduce redundant information, in contrast to existing approaches that utilize all caption data in an external memory bank. Additionally, in contrast to traditional single-layer memory structures that read information based on similarity, we utilized a hierarchical clustering method to construct a hierarchical memory structure. This allows for the retrieval of information in a stepwise manner, starting from abstract to more detailed information.

% \begin{figure*}[t]
% \centering
% \vfill
% \mpage{0.49}{
%     \includegraphics[width=1\textwidth]{AnonymousSubmission/LaTeX/figure/1_Overall_architecture.pdf}\
% }
% \hfill
% \mpage{0.49}{
%     \includegraphics[width=1\textwidth]{AnonymousSubmission/LaTeX/figure/3_read.pdf}
% }

% \mpage{0.49}{
%     (a) Overall Architecture
% }
% \hfill
% \mpage{0.49}{
%     (b) Hierarchical Memory Read
% }
% \vspace{-0.2cm}
% \caption{Overview of ours.}
% \label{fig:overview}
% \end{figure*}

\begin{figure*}[t]
\centering
\vfill
\mpage{0.44}{
    \includegraphics[width=1\textwidth]{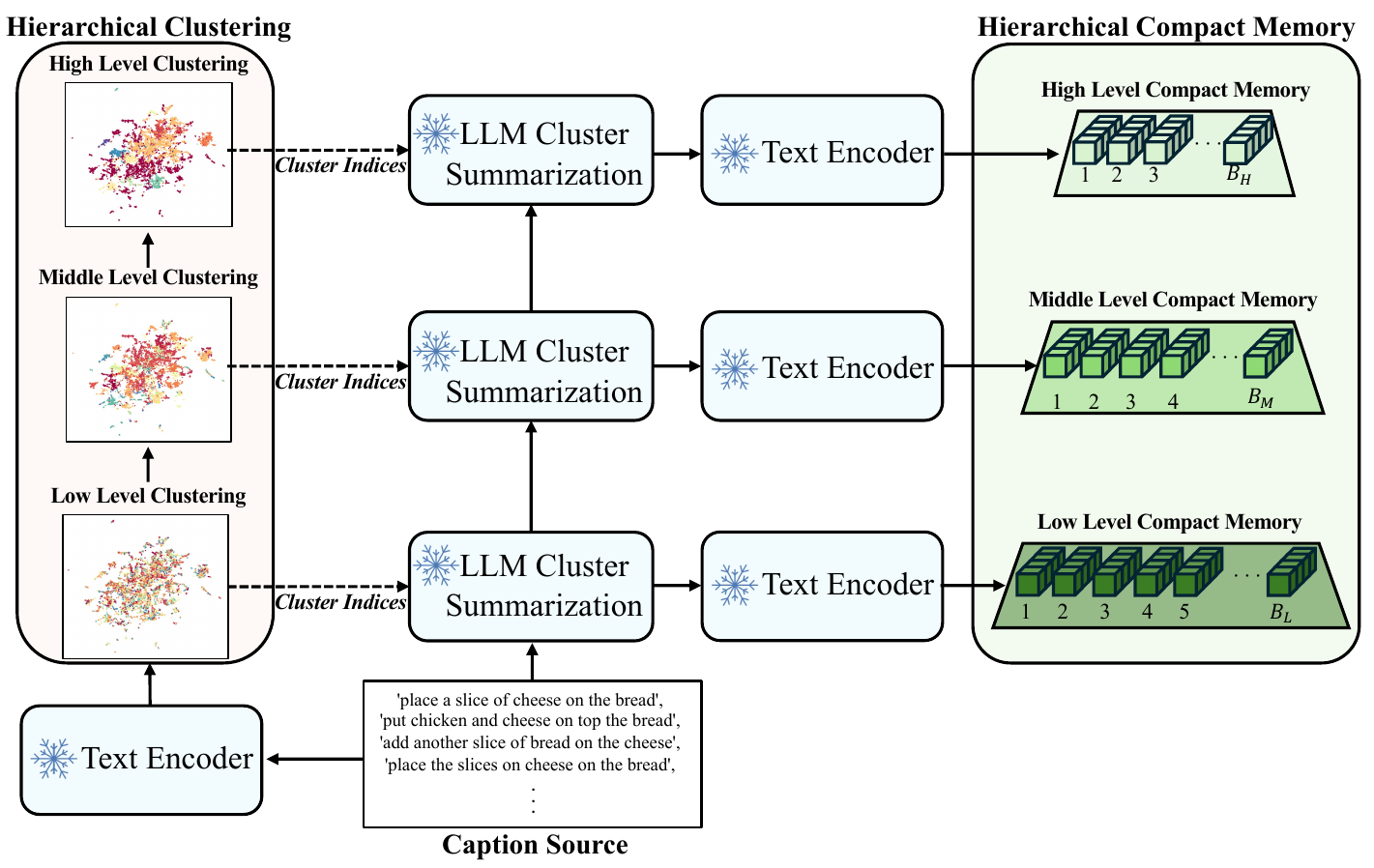}\
}
\hfill
\mpage{0.5}{
    \includegraphics[width=1\textwidth]{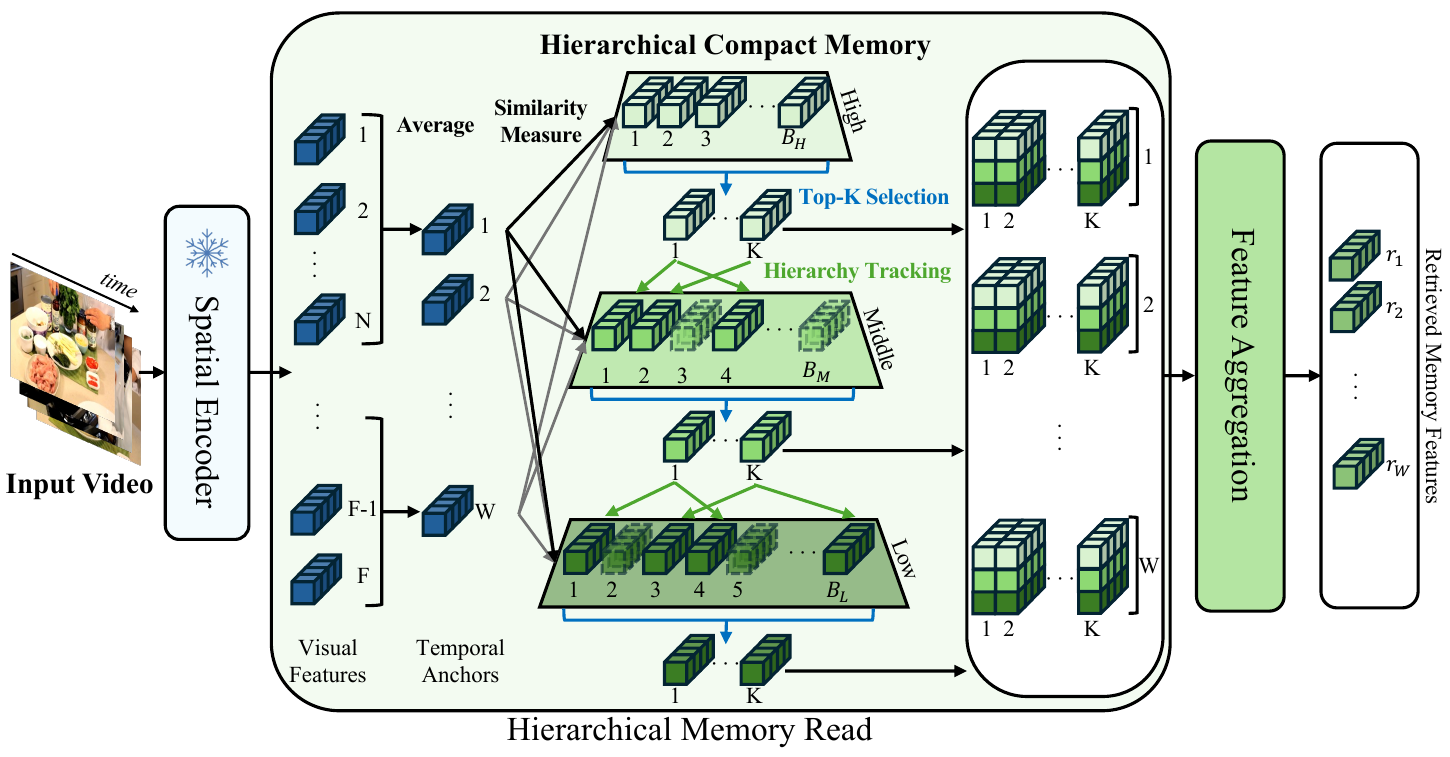}
}

\mpage{0.49}{
    (a) Hierarchical Memory Construction
}
\hfill
\mpage{0.49}{
    (b) Hierarchical Memory Read
}
% \vspace{-0.2cm}
% \vspace{-0.2cm}
\caption{\textbf{The construction and read process of hierarchical compact memory.} %We illustrate the construction and retrieval process of hierarchical compact memory.
As illustrated in (a), hierarchical memory is constructed using iterative clustering in a bottom-up approach to recall relevant episodes and abstract concepts, with LLM-based summarization generating compact, memory-efficient representations. As illustrated in (b), We retrieve a segmented visual cue from each temporal anchor, beginning at the high-level, abstract layer, and then recursively exploring lower layers for additional relevant information. K features are retrieved per level, and this process is repeated for all temporal anchors.}
\label{fig:Hi_Mem}
\end{figure*}

\section{Method}
\label{sec:Method}

The goal of this study is to enhance event localization and caption generation in untrimmed videos by leveraging prior knowledge. We introduce HiCM$^2$, a novel framework that utilizes cross-modal retrieval from a hierarchical compact memory. To achieve this, we construct the memory using a hierarchical organization combined with compact representations, enabling the effective recall of relevant episodes and abstract concepts while minimizing redundancy (Section \ref{sec:sub_mem_con}). Additionally, we propose a method for effective hierarchical memory retrieval. By employing a top-down approach, we begin with high-level abstract information and gradually access finer details from lower levels. This ensures the effective extraction of meaningful information from the hierarchical memory, improving both event localization and caption generation in DVC (Section \ref{sec:sub_ret}). 

Figure \ref{fig:overall} shows the overall architecture of our method.
For the given input video, the model generates segment and caption pairs \(\{(T^S_n,T^E_n,S_n)\}^{N}_{n=1}\) where $N$ denotes the number of events detected by our method and $T^S_n$ and $T^E_n$ denote the start and the end timestamp of $n$-th event. $S_n$ denotes the generated captions for $n$-th event segment. Details of dense event prediction will be introduced in Section~\ref{sec:denseprediction}.

\subsection{Hierarchical Compact Memory Modeling}
\label{sec:sub_mem_con}
In this section, we describe the method for constructing a hierarchical memory and generating compact representations. To recall relevant episodes and abstract concepts, we build a hierarchical memory using iterative clustering in a bottom-up approach. To minimize redundancy while creating a memory-efficient compact structure, we generate compact representations through LLM-based summarization.

To ensure the memory bank stores high-quality semantic information, we use captions from the training dataset. For segment-level video-to-text retrieval, we utilize the pre-trained CLIP ViT-L/14 model \cite{dosovitskiy2020image, radford2021learning}, which effectively aligns images and text within a shared embedding space.

\subsubsection{Building Hierarchical Memory.}
To construct a hierarchical memory structure, we leverage the hierarchical clustering method FINCH \cite{sarfraz2019efficient}, which performs iterative clustering using a bottom-up approach. FINCH forms clusters based on the observation that the first neighbor of each data point is sufficient to discover a connected chain among data points. Furthermore, as a hierarchical clustering method, FINCH generates a meaningful clustering tree that interprets data at different levels of granularity, making it an appropriate approach for constructing hierarchical memory. Note that the number of memory levels is automatically determined based on the FINCH clustering results for each dataset.

As illustrated in Figure \ref{fig:Hi_Mem} (a), we first encode the caption sources using the CLIP text encoder. Then, we perform iterative clustering on the encoded sentences until only abstract clusters that can no longer be grouped remain, resulting in a hierarchical clustering structure. Since each level's clusters are formed by the connections between clusters from the lower level, we obtain clustering indices for each level. These indices are then used to construct the memory components at each hierarchical level. With this, we construct hierarchical memory that contains high-level abstract information and finer details from lower levels.

\subsubsection{Compact Representation with LLM.}
To minimize redundancy and effectively recall information from memory, we store compact representations generated by summarization using an LLM. For this purpose, we adopt the pre-trained Llama3 70B model \cite{dubey2024llama}. We use the LLM to store representative information derived from clusters. We provide the LLM with instructions for summarizing clustered sentences from each hierarchical level. This process results in the creation of a summarized sentence-level feature for each cluster, effectively removing outliers and capturing core semantic information within the cluster. Additionally, through summarization, we reduce approximately 10K memory components to around 2K, constructing a memory-efficient compact structure.

As in Figure \ref{fig:Hi_Mem} (a), we first summarize the sentences of clusters from the caption source using clustered indices for the lower level. From this, we obtain one summarized sentence per cluster. Consequently, at each memory level, an encoded memory component is created for each cluster, thereby obtaining compact representations. Then, the summarized sentences from this level serve as input for the next level's summarization, where they are further summarized according to the next level's cluster indices. We repeat this process across all levels of the memory, constructing a compact memory structure that facilitates efficient recall.

\subsection{Retrieval from Hierarchical Memory}
\label{sec:sub_ret}

The hierarchical compact memory is structured with a high level containing abstract information and lower levels that provide more detailed information. Consequently, to effectively utilize this well-composed hierarchical compact memory, it is important to systematically search through the levels, moving from abstract to detailed, to retrieve and utilize information in an appropriate manner. For this, we propose a retrieval method that begins by searching for information related to the visual cue within the most abstract, high-level layer of the hierarchical memory bank and then recursively explores lower layers for additional relevant information. 

As illustrated in Figure \ref{fig:Hi_Mem} (b), we divide the input video into $W$ temporal anchors to extract semantic information related to events, as untrimmed videos may contain multiple events. We use CLIP ViT-L/14 for frame-level visual feature extraction and compress these features by averaging within each anchor, resulting in segment-level visual features. These features then serve as queries to retrieve relevant information from the external memory for each anchor. We then calculate the similarity between the segment-level visual cues and the memory features in the high-level layer, selecting the most similar features. Subsequently, the corresponding clusters containing these selected features are tracked down to the lower layers, where we calculate the similarity between the visual cue and the more detailed clusters that result from the merging process. This recursive process continues across the lower layers to obtain increasingly specific information. For the relevant features obtained, we perform feature aggregation through two rounds of average pooling. First, we conduct average pooling on the most similar features selected at each level to create a single representative feature. Then, we take one feature from each level and perform average pooling again to integrate the information from all levels. This entire process is repeated for each of the $W$ temporal anchors. Finally, we obtain the retrieved text features \(\mathbf{r}=\{r_a\}^W_{a=1}\).

\subsection{Dense Event Prediction}
\label{sec:denseprediction}

\subsubsection{Architecture.}
This section describes a framework of structure that integrates visual features with retrieved text features to enhance event localization and captioning.

As illustrated in Figure \ref{fig:overall}, the visual encoder processes a sequence of \(F\) frames. The spatial encoder outputs embeddings for each frame, which are then fed into a temporal encoder composed of transformer blocks to account for temporal interactions between frames. This process yields the final temporal visual features \(\mathbf{\tilde{x}} = \{\tilde{x}_i\}_{i=1}^F\) as the output of the visual encoder. We use CLIP ViT-L/14 as the backbone to extract spatial embeddings from the input frames, keeping it in a frozen state for ease of use. 

For memory retrieval, the spatial embeddings extracted by the visual encoder are used to generate segment-level temporal anchors. The embeddings for each frame are divided into $W$ temporal anchors, which serve as queries to the hierarchical memory bank. Each anchor retrieves relevant hierarchical information from the memory based on similarity. Then, we get \(\mathbf{r}=\{r_a\}^W_{a=1}\). 
We also use speech information. For the text of speech transcript, the encoder's token embedder independently embeds each token, producing semantic embeddings. For the given time information of speech, we begin with a text tokenizer that has a vocabulary size of \( v \), to which we add \( b \) time tokens, thus expanding the tokenizer to encompass \( v + b \) tokens. These time tokens denote relative timestamps within a video, which we achieve by dividing a video into \( b \) evenly spaced timestamps. Specifically, we employ the SentencePiece tokenizer~\cite{kudo2018sentencepiece}, which features a vocabulary size of \( v = 32,128 \) and includes \( b = 100 \) time tokens.
Subsequently, the transformer-based text encoder calculates the interactions between the tokens of the speech transcripts, outputting text features \(\mathbf{y}=\{y_j\}^s_{j=1}\). The text decoder uses the retrieved memory features, temporal visual features, and text features that compose the encoder embeddings to autoregressively generate a target event sequence consisting of segment and caption pairs.

\subsubsection{Training and Inference.}
During training, we train HiCM$^2$ by minimizing the following loss function: 

{\small
\begin{equation}
\mathcal{L}_{\theta}(x, y, r, z) = -\frac{1}{\sum_{c=1}^{L-1} w_c} \sum_{c=1}^{L-1} w_c \log p_{\theta}(z_{c+1} \mid x, y, r, z_{1:c}),
\end{equation}
}

\noindent where $L$ represents the length of the target sequence in the decoder, $w_c$ indicates the weight assigned to the c-th token in the sequence, uniformly set to $w_c = 1 \ \forall c$ in practice, $\theta$ symbolizes the trainable parameters within the model, and $p_{\theta}$ is the probability distribution generated by the model over the vocabulary of text and time tokens. Our approach, based on the sequence-to-sequence structure of the Vid2Seq model \cite{yang2023vid2seq}, fine-tunes a model pre-trained on approximately 1.8 million videos \cite{yang2023vidchapters}.
During inference, given visual features $x$, given text features $y$, and retrieved text input $r$, our model predicts a target event sequence $z$.
For both training and inference, we conducted retrieval using the same hierarchical compact memory bank.

% \begin{table}[]
\begin{table}
\centering
\resizebox{0.99\columnwidth}{!}{
    \begin{tabular}{c|c|cccc}
    \toprule
    \multirow{1}{*}{Memory Construction} & Hierarachy &CIDEr & METEOR & SODA$\_c$& F1 \\
    \midrule
    No Memory     & \xmark&66.29 & 12.41 & 9.87  & 31.08 \\

    All Training Captions& \xmark   &  67.90   & 12.49   &  10.38  &  32.31  \\ %Bleu4:5.73
    \midrule

    Clustering    & \cmark  &   67.15   &  \textbf{ 12.97}  &   10.17 &32.30   \\
    Clustering+LLM(Ours) & \cmark & \textbf{71.84} & 12.80 & \textbf{10.73} & \textbf{32.51} \\ 
    \bottomrule
    \end{tabular}
}
\caption{\textbf{Effect of hierarchical compact memory retrieval on YouCook2.}}
\label{tab:construction_abl}
\end{table}

% \begin{table*}[]
\begin{table*}
\centering
\scalebox{0.9}{
\begin{tabular}{@{}l|c|cccc|cccc@{}}
\toprule
\multirow{2}{*}{Method}&\multirow{2}{*}{PT} & \multicolumn{4}{|c}{YouCook2(val)} & \multicolumn{4}{|c}{ViTT(test)} \\
&&CIDEr& METEOR & SODA$\_c$ & BLEU4 &CIDEr& METEOR & SODA$\_c$ & BLEU4 \\
\midrule
PDVC          & \xmark      & 29.69 & 5.56  & 4.92 &1.40 &  -   &  -    &     -   &-\\
CM$^{2}$      & \xmark  & 31.66 & 6.08  & 5.34 & 1.63&  -   &  -    &  -     & -\\
\midrule
Streaming V2S & \cmark   & 32.90 & 7.10  & 6.00 &- & 25.2 & 5.80& 10.00   & - \\
DIBS          & \cmark & 44.44 & 7.51  & 6.39 &- &   -  &   -   &  -   &  -\\
Vid2Seq{$^{\dagger}$} &\cmark & \underline{66.29} & \underline{12.41} & \underline{9.87} &\underline{5.64} & \underline{48.84} &\underline{9.51}& \underline{14.99}  & \underline{0.71} \\
HiCM$^2$(Ours)          &\cmark & \textbf{71.84} & \textbf{12.80} & \textbf{10.73} &\textbf{6.11} & \textbf{51.29}&\textbf{9.66}&\textbf{15.07} &\textbf{0.86}\\ 
\bottomrule
\end{tabular}
}
\caption{\textbf{Performance of event captioning.}
        Bold means the highest score. PT denotes pretraining from the additional video datasets. {$^{\dagger}$} denotes results reproduced from official implementation in our environment. All methods use CLIP backbone. 
    }
\label{tab:SOTA_cap_tab}
\end{table*}

%%%%%%%%%%%%%%%%%%%%%%%
\section{Experiments}
\label{sec:Experiments}
To evaluate the effectiveness of our method, we conducted comparative experiments. Section~\ref{subsection:settings} describes the experimental setup used in this study. Section~\ref{subsection:hier} highlights the role of memory retrieval in DVC. Section~\ref{subsection:sota} presents a comparison with state-of-the-art methods. 
Section~\ref{subsection:disc} provides ablation studies to validate the contribution of each component in our model. We also provide qualitative results.

\subsection{Experimental Settings}
\label{subsection:settings}

\subsubsection{Dataset.}
We utilized two DVC benchmark datasets, YouCook2~\cite{zhou2018towards} and ViTT \cite{huang2020multimodal}, for both training and evaluation. The YouCook2 dataset includes 2,000 untrimmed videos depicting cooking procedures, with an average duration of 320 seconds per video and 7.7 temporally localized sentences per annotation. We employed the standard dataset split for training, validation, and testing purposes. The ViTT dataset includes 8,000 untrimmed instructional videos, each with an average length of 250 seconds and annotated with 7.1 temporally localized short tags. It is worth noting that we utilized approximately 10\%$\sim$20\% fewer videos than the original dataset, as we only included those accessible on YouTube~\cite{yang2023vid2seq}.
\subsubsection{Evaluation Metrics.}
We evaluated our method on two sub-tasks in DVC. By using the official evaluation tool~\cite{wang2020dense}, we measured the quality of the generated captions with the CIDEr~\cite{vedantam2015cider}, BLEU4~\cite{papineni2002bleu}, and METEOR~\cite{banerjee2005meteor} metrics, which compare the generated captions to the ground truth across IoU thresholds of {0.3, 0.5, 0.7, 0.9}. Additionally, to assess storytelling ability, we used the SODA\_c metric~\cite{fujita2020soda}. For event localization, we calculated the average precision, average recall, and F1 score (the harmonic mean of precision and recall), averaging these metrics over IoU thresholds of {0.3, 0.5, 0.7, 0.9}.
\subsubsection{Implementation Details.}
For both datasets, video frames were extracted at a rate of 1 frame per second. The sequences were then either subsampled or padded to ensure a total of $F$ frames, where $F$ was set to 100. 
Both the text encoder and decoder are initialized using a pretrained T5-Base model \cite{raffel2020exploring}. 

The hierarchical memory is organized into 4 levels for YouCook2 and 5 layers for ViTT. For YouCook2, we use 10,337 sentences to construct compact memory across levels, with \{1758, 313, 68, 8\} memory units allocated per level. For ViTT, we utilize 34,599 sentences, distributing compact memory across levels as follows: \{4487, 741, 114, 12, 3\} memory units per level. For YouCook2, the number of anchors is set to 10 and the number of retrieved features is set to 10 for each anchor level. For ViTT, the number of anchors is set to 30 and the number of retrieved features is set to 30 for each anchor level.
\subsection{Effect of Hierarchical Memory Retrieval}
\label{subsection:hier}

To evaluate the effectiveness of memory retrieval with hierarchical compact memory, we conducted comparative experiments on YouCook2. As in Table~\ref{tab:construction_abl}, we compare four different memory retrieval methods. ``No Memory" refers to the test results of an encoder-decoder structure without memory retrieval. ``All Training Captions" involves storing all memory captions as features in a flat memory structure without any hierarchical organization. ``Clustering" utilizes the FINCH algorithm~\cite{sarfraz2019efficient} to create a hierarchical structure, using averaged features of clustered sentences as memory elements. In this method, a bottom-up approach is applied, where layers above the lowest level are formed by averaging features from the previous level, based on clustering result indices. ``Clustering + LLM" involves generating representative captions for each cluster through summarization with LLM, which are then used as memory elements. This approach, implemented in a bottom-up fashion, represents the hierarchical compact memory. 

As shown in Table~\ref{tab:construction_abl}, the results demonstrate that storing and retrieving all training captions in memory leads to a performance improvement compared to using no memory at all. Further enhancements are observed when employing hierarchical clustering to create a structured memory, with additional performance gains achieved by generating prototypes through LLM-based summarization. This suggests that while average pooling after hierarchical clustering provides some benefit, the use of LLMs for summarization is more effective in creating semantic prototypes, resulting in more substantial performance improvements.
%, particularly in handling outliers, resulting in more substantial performance improvements.

\subsection{Comparison with State-of-the-art-Methods}
\label{subsection:sota}

In Table~\ref{tab:SOTA_cap_tab}, we compare our method with state-of-the-art approaches~\cite{wu2024dibs,zhou2024streaming,yang2023vid2seq} on both the YouCook2 and ViTT datasets. As shown in Table \ref{tab:SOTA_cap_tab}, the methods are categorized based on their use of pretraining, with performance measured across several key metrics, including CIDEr, METEOR, SODA$_c$, and BLEU4. The results indicate that methods employing pretraining generally outperform those that do not, highlighting the importance of pretraining in improving caption quality. Our method achieves the best scores across the four metrics among the models that use pre-trained knowledge. By using prior knowledge from external hierarchical memory, our method improves the quality of caption generation across all metrics on both datasets. This performance advantage underscores the effectiveness of our approach in generating accurate and contextually appropriate captions.

\begin{table}
% \begin{table}[]
\centering
\scalebox{0.72}{
\begin{tabular}{@{}l|c|ccc|ccc@{}}
\toprule
\multirow{2}{*}{Method}&\multirow{2}{*}{PT} & \multicolumn{3}{|c}{YouCook2(val)} & \multicolumn{3}{|c}{ViTT(test)} \\
&&F1& Recall & Precision & F1&Recall & Precision \\
\midrule
    PDVC          & \xmark                & 26.81 & 22.89 & 32.37&  - &  - & - \\
CM$^{2}$      &\xmark     & 28.43 & 24.76& \underline{33.38} &  - & -  & - \\
\midrule
Streaming V2S & \cmark             & 24.10 & - &-  & 35.40  &  - & - \\
DIBS          & \cmark          & \underline{31.43} & 26.24 & \textbf{39.81}& -  & -  & - \\
Vid2Seq{$^{\dagger}$} & \cmark    & 31.08 &\underline{30.38} & 31.81& \textbf{46.21}  & \textbf{45.89}  & \underline{46.53}  \\
HiCM$^2$(Ours)          & \cmark & \textbf{32.51} &\textbf{ 32.51} & 32.51&\underline{45.98}&\underline{45.00} & \textbf{47.00}  \\
\bottomrule
\end{tabular}
}
\caption{\textbf{Performance of event localization.}
        Bold means the highest score. {$^{\dagger}$} denotes results reproduced from official implementation in our environment. %All methods use CLIP backbone. 
    }
\label{tab:SOTA_loc_tab}
\end{table}

In Table~\ref{tab:SOTA_loc_tab}, we compare the performance of various methods on the event localization task across the YouCook2 and ViTT datasets. The methods are evaluated based on whether they utilize pretraining, with key metrics including F1, Recall, and Precision. Our method, HiCM$^2$, demonstrates comparable performance across both datasets. HiCM$^2$ effectively leverages hierarchical compact memory, resulting in more accurate event localization. The results highlight the advantages of our approach in capturing complex temporal patterns within untrimmed videos. These findings underscore the robustness and effectiveness of HiCM$^2$ in enhancing event localization tasks. 
The results from both captioning and localization performance indicate that pre-trained prior knowledge and retrieval-augmented prior knowledge complement each other, creating a synergistic effect.
\subsection{Discussion}
% \label{subsection:abl}
\label{subsection:disc}

\begin{table}
% \begin{table}[]
\scalebox{0.88}{
\centering
\begin{tabular}{@{}ccc|ccccc@{}}
\toprule
High & Middle & Low & CIDEr & METEOR&  SODA$\_c$ &F1 \\
%Encoding & Cross-Attention & &&& \\
\midrule

% \xmark & \xmark& \xmark & 66.29 & 12.41 &  9.87 & 31.08 \\
% \xmark & \xmark& \cmark & 67.75 & 12.33 &  10.35 & 32.27 \\
% \xmark&\cmark & \xmark & 67.59  & 12.45 & 10.37 & 32.23 \\ %B4:5.59
% \cmark & \xmark& \xmark & 67.21  & 12.21 & 10.28 & 31.98 \\
% \xmark &\cmark & \cmark & 68.61  &  12.35 & 10.51 & 32.07 \\ %b4:5.88
% \cmark &\xmark & \cmark & 69.81  &  12.63 & 10.54 &\textbf{33.00} \\
% \cmark &\cmark & \xmark &68.05  & 12.33 & 10.41 & 32.87 \\ %B4:5.97
% \cmark &\cmark & \cmark & \textbf{71.84} & \textbf{12.80} & \textbf{10.73} & 32.51 \\

 & &  & 66.29 & 12.41 &  9.87 & 31.08 \\
& & \cmark & 67.75 & 12.33 &  10.35 & 32.27 \\
&\cmark &  & 67.59  & 12.45 & 10.37 & 32.23 \\ %B4:5.59
\cmark & && 67.21  & 12.21 & 10.28 & 31.98 \\
 &\cmark & \cmark & 68.61  &  12.35 & 10.51 & 32.07 \\ %b4:5.88
\cmark & & \cmark & 69.81  &  12.63 & 10.54 &\textbf{33.00} \\
\cmark &\cmark &  &68.05  & 12.33 & 10.41 & 32.87 \\ %B4:5.97
\cmark &\cmark & \cmark & \textbf{71.84} & \textbf{12.80} & \textbf{10.73} & 32.51 \\
\bottomrule
\end{tabular}
}
\caption{\textbf{The ablation study for the use of hierarchical levels memory.} Among the four hierarchical levels generated from the YouCook2 training data, the high level corresponds to level 4, the middle level includes levels 2 and 3, and the low level corresponds to level 1.}
% \vspace{-0.2cm}
\label{tab:level_abl}
\end{table}

\begin{figure*}[t]
    \centering
    \includegraphics[width=0.8\linewidth]{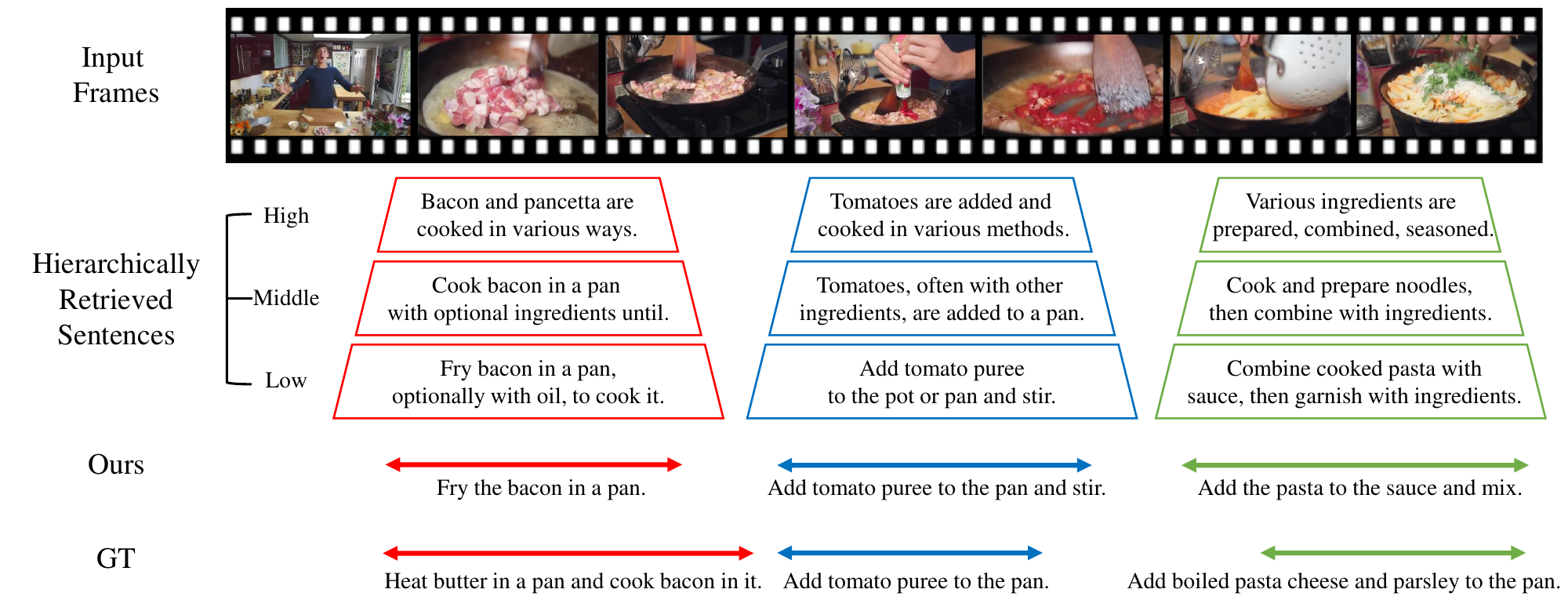}
    \caption{\textbf{Example of predictions on the YouCook2 Validation set using our approach.} The hierarchical retrieved sentences shown are examples of retrieval results with the highest semantic similarity at each hierarchical level, corresponding to specific segments of the input frames. Each retrieved sentence is converted into features and utilized in our model's predictions for the segments. Matching colors indicate the association between retrieved knowledge and prediction.}
    \label{fig:qualitative}
\end{figure*}

\subsubsection{Effect of Each Level in Hierarchy.}
Table \ref{tab:level_abl} presents an ablation study evaluating the impact of utilizing hierarchical memory at different levels on performance. The high level is the level that contains the most abstract concepts. The middle level contains more detailed actions and situations, providing a balance between abstract concepts and specific details. The low level focuses on concrete details and events accurately describing specific actions, objects, and events within the scenes. The results clearly demonstrate the importance of incorporating memory across all hierarchical levels, as the best performance is achieved when all three levels (high, middle, and low) are utilized together.

We show the effectiveness of a comprehensive hierarchical memory approach, where each level contributes uniquely to improving the overall performance.

\subsubsection{Qualitative Result.}

Figure~\ref{fig:qualitative} shows an example of predictions by our approach, demonstrating effective use of memory retrieval that references meaningful and helpful sentences from memory across different levels, achieved through segment-level video-text retrieval for the given video. As shown in the figure, sentences formed at a high level typically include words that encompass a variety of methods or ingredients. At the middle level, while still abstract about various ingredients, there is a more detailed description of actions and contexts. At the low level, sentences precisely describe the specific subjects and actions appearing in the scene. As a result, our method generates relatively accurate event boundaries and captions with abstract concepts and detailed descriptions. 
\subsubsection{Effect of Number of Retrieved Features for Each Level.}

\begin{table}
% \begin{table}[]
\centering
\scalebox{0.95}{
\begin{tabular}{c|cccc}
\toprule
\multirow{1}{*}{\#K} &  CIDEr & METEOR & SODA$\_c$& F1 \\
\midrule
1   &    67.47 &    12.39     &  10.53      &  32.27   \\
5   & 68.15 & 12.58  & 10.33 & \textbf{32.51} \\
10  & \textbf{71.84} & \textbf{12.80} & \textbf{10.73} & \textbf{32.51}\\
20 & 67.97&12.22&10.40&31.56\\ 
\bottomrule
\end{tabular}
}
\caption{\textbf{Effect of the number of selected features in YouCook2.} \#K denotes the number of retrieved text features per anchor from the memory bank at each level.}

\label{tab:topk_abl}
\end{table}
We also explored the effect of the number of retrieved features per anchor as shown in Table~\ref{tab:topk_abl}.
Setting this number to 1 means that only the most similar text from the memory, relative to the visual feature of the temporal anchor, is considered. Increasing this number generally provides the model with more stable and robust semantic information. When the number of retrieved features per anchor reaches 10, our approach consistently delivers strong performance across both sub-tasks in the YouCook2 dataset. Note that the number of retrieved features has been uniformly applied across all levels. However, retrieving too many features can introduce noise, as they are selected based on descending similarity, potentially lowering the overall performance.

%%%%%%%%%%%%%%%%%%%%%%%
\section{Conclusion}
\label{sec:Conclusion}

In this study, we introduce a novel DVC model that leverages hierarchical compact memory with cross-modal retrieval, effectively mimicking human memory processes for improved event localization and description. By organizing memory from abstract to detailed information and employing a top-down retrieval process, the model ensures precise and coherent information retrieval. Through comprehensive experiments on the YouCook2 and ViTT datasets, we demonstrate the effectiveness of our hierarchical memory retrieval approach. Notably, HiCM$^2$ achieved state-of-the-art performance on both datasets. Our work highlights the potential of memory-augmented models to enhance DVC. Additionally, our findings suggest that the synergy between pre-trained prior knowledge and retrieval-augmented knowledge in our approach could complement existing pretraining efforts in vision-and-language tasks, potentially offering a promising avenue for further improvement in the field.

\section*{Acknowledgments}
This work was supported in part by the Institute of Information and Communications Technology Planning and Evaluation (IITP) Grant funded by the Korea Government (MSIT) under Grant 2020-0-00004 (Development of Provisional Intelligence Based on Long-term Visual Memory Network), Grant 2022-0-00078, Grant IITP-2024-RS-2023-00258649, Grant RS-2022-00155911, Grant 2021-0-02068, by the National Research Foundation of Korea (NRF) Grant funded by the Korea Government (MSIT) under Grant RS-2024-00334321, and by Center for Applied Research in Artificial Intelligence (CARAI) grant funded by DAPA and ADD (UD230017TD).

\bibliography{aaai25}
% \clearpage
% \setcounter{page}{1}
% \maketitlesupplementary
% \appendix
% \twocolumn[
% \begin{@twocolumnfalse}
% \begin{center}
% \textbf{\LARGE Appendix}
% \vspace{2em}
% \end{center}
% \end{@twocolumnfalse}
% ]
% \begin{appendices}
\twocolumn[{
    \begin{center}
        {\LARGE \textbf{Supplementary Material}}\\[1em]
        {\LARGE \textbf{HiCM$^2$: Hierarchical Compact Memory Modeling for Dense Video Captioning}}
        \vspace{1em}
    \end{center}
}]

\appendix

\section{Additional Analysis of HiCM$^2$}

\begin{table}
% \begin{table}[]
\centering
\begin{tabular}{c|cccc}
\toprule
Memory &\multirow{2}{*}{ CIDEr} & \multirow{2}{*}{METEOR }& \multirow{2}{*}{SODA\_c}&\multirow{2}{*}{ F1} \\
Selection & &  & & \\
\midrule
Max      &   67.47 &    12.39     &  10.53      &  32.27  \\
Top-K    & \textbf{71.84} & \textbf{12.80} & \textbf{10.73} & \textbf{32.51} \\
Similarity &      67.77  &   12.31     &  10.29   &   32.40 \\
\bottomrule
\end{tabular}
\caption{\textbf{Effect of memory selection types in YouCook2.} Max refers to selecting the single retrieved feature with the highest similarity from each level, Top-K refers to selecting the top K retrieved features with the highest similarity from each level, and Similarity refers to selecting retrieved features from each level that exceed a defined similarity threshold.
}
\label{tab:selection_abl}
\end{table}
\subsection{Effect of Memory Selection.}
We compare different memory retrieval methods on the performance in Table \ref{tab:selection_abl}. The results indicate that the Top-K method, which retrieves and averages multiple memory entries, outperforms the other methods across all metrics. This suggests that accessing a broader range of relevant information through the Top-K approach leads to more robust and contextually accurate captions, as well as improved event localization. These findings underscore the effectiveness of the Top-K retrieval strategy in enhancing overall model performance.

\begin{table}[t!]

\centering
\resizebox{0.92\columnwidth}{!}{
    \begin{tabular}{@{}c|cccc@{}}
    \toprule
    \multirow{1}{*}{\# Anchor} & CIDEr & METEOR  &SODA$\_c$& F1 \\
    \midrule
    1 & 67.95	&12.41&	10.14&	31.65\\
    5 & 69.87&	12.59&	10.58&	32.34\\
    10 & \textbf{71.84} & \textbf{12.80} & \textbf{10.73} & \textbf{32.51}\\
    20 & 66.37	&12.35	&10.10&	32.49 \\
    \bottomrule
    \end{tabular}
}
\vspace{-0.2cm}
\caption{\textbf{Effect of anchor number for retrieval in YouCook2.} \# Anchor denotes the number of anchors. The performance is measured by changing the number of anchors.
}
\label{tab:anchor_num}
\end{table}

\subsection{Effect of Anchor Number for Retrieval}
We investigate the impact of varying the number of temporal anchors during memory retrieval. The temporal anchors determine the basic units for querying the memory bank and influence the number of retrieved features. Table \ref{tab:anchor_num} presents the performance results as the anchor number changes on the YouCook2 dataset. With only one anchor, the entire untrimmed video is averaged into a single visual feature for querying, which limits the ability to capture fine-grained details necessary for retrieving precise semantic text cues. Increasing the number of anchors allows the model to capture more detailed information, improving performance in dense video captioning. However, retrieving information from too many temporal segments can introduce noise, leading to a decline in performance. The results show that setting the anchor number to 10 consistently delivers strong performance in both event localization and caption generation on the YouCook2 dataset.

\begin{table}[]
\centering
\begin{tabular}{c|cccc}
\toprule
Retrieval &\multirow{2}{*}{ CIDEr} & \multirow{2}{*}{METEOR }& \multirow{2}{*}{SODA\_c}&\multirow{2}{*}{ F1} \\
Pass & &  & & \\
\midrule
Concat    & \textbf{71.84} & \textbf{12.80} & \textbf{10.73} & \textbf{32.51} \\
Cross Attn      & 69.90	&12.38	&10.37&	31.37\\
\bottomrule
\end{tabular}
\caption{\textbf{Effect of retrieval pass types in YouCook2.} Concat refers to the method of passing retrieved text features to the text decoder by concatenating them with visual and speech features. Cross Attn refers to the method of adding a new cross-attention layer to the text decoder for the retrieved text features.}
\label{tab:ret_pass}
\end{table}

\subsection{Effect of Retrieval Pass Types}
We investigate the impact of different methods for passing retrieved text features to the text decoder. Specifically, we compare two approaches: concatenating the retrieved text features with visual and speech features before feeding them into the text decoder and adding a new cross-attention layer in the text decoder specifically for the retrieved text features. In table \ref{tab:ret_pass}, the results demonstrate that the Concat method outperforms the Cross Attn approach across all evaluation metrics, indicating that directly merging the retrieved text features with other modalities is more effective for this sequence prediction structure.

\begin{table}[]
\centering
\begin{tabular}{c|cccc}
\toprule
Hierarchical &\multirow{2}{*}{ CIDEr} & \multirow{2}{*}{METEOR }& \multirow{2}{*}{SODA\_c}&\multirow{2}{*}{ F1} \\
Aggregation & &  & & \\
\midrule
\xmark      & 69.67&	12.74	&10.37&	\textbf{33.43}\\
\cmark    &\textbf{ 71.84 }& \textbf{12.80} & \textbf{10.73} & 32.51 \\
\bottomrule
\end{tabular}
\caption{\textbf{Effect of hierarchical aggregation in YouCook2.} Hierarchical Aggregation refers to the aggregation of retrieved text features across different levels in the Hierarchical Memory Read process.
}
\label{tab:hier_aggregation}
\end{table}

\subsection{Effect of Hierarchical Aggregation}
We investigate the impact of hierarchical aggregation on performance. Specifically, we examine the differences between applying hierarchical aggregation, where retrieved text features are combined across different levels in the Hierarchical Memory Read process, and not applying it. As shown in Table \ref{tab:hier_aggregation}, the results indicate that enabling hierarchical aggregation leads to improvements across all evaluation metrics, suggesting that aggregating information from multiple levels enhances the model’s ability to generate more accurate and contextually appropriate captions. Conversely, not performing hierarchical aggregation enhances localization ability.

\begin{figure*}[t]
    \centering
    \includegraphics[width=0.7\linewidth]{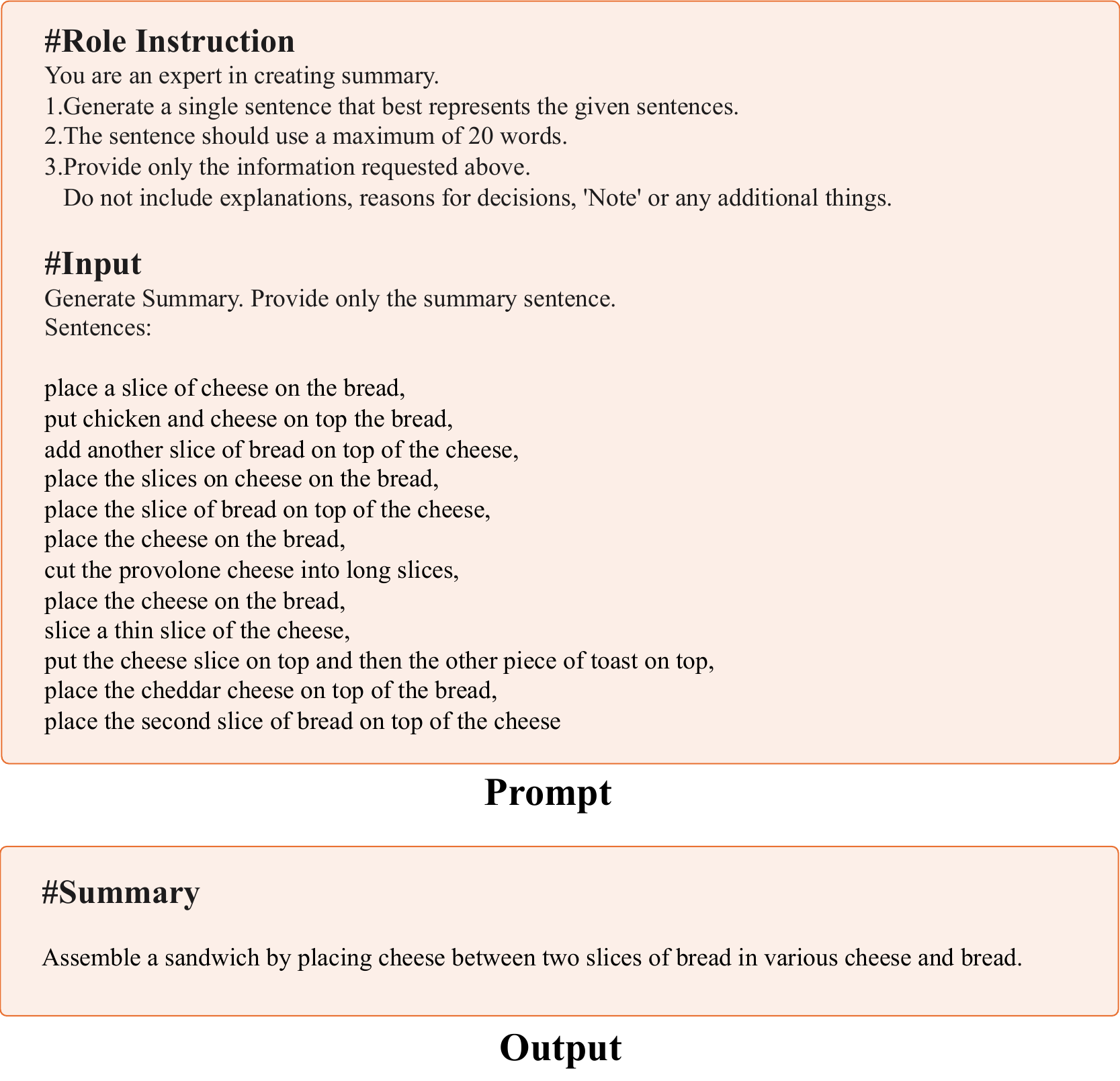}
    \caption{\textbf{Example of LLM summarization instruction on the YouCook2 training set on our approach.}}
    \label{fig:supp_instruction}
\end{figure*}

\begin{table}[]
\centering
\scalebox{0.91}{
\begin{tabular}{c|c|cccc}
\toprule
\multirow{1}{*}{LLM} & \#Params& CIDEr & METEOR & SODA$\_c$& F1 \\
\midrule
Llama3& 8B   & 68.95&	12.31&	10.49&	32.51\\
Llama3 &70B  & 71.84 & 12.80& 10.73 &32.51\\
\bottomrule
\end{tabular}
}
\caption{\textbf{Effect of different LLM models for memory summarization on YouCook2.} \#Params denotes the number of parameters in the LLM.}

\label{tab:LLAMA}
\end{table}
\subsection{Effect of LLM Models for Compact Memory }
In Table \ref{tab:LLAMA}, the results show that the larger Llama3 model generally improves the quality of the generated summaries, leading to a better compact memory, which in turn enhances performance in dense video captioning, particularly in metrics such as CIDEr and METEOR. This suggests that a more powerful LLM enhances the semantic richness and relevance of the summaries used in the hierarchical compact memory construction, contributing to more accurate and contextually appropriate summarized captions.

\section{Implementation Detail}
\subsection{Instruction of LLM Summarization}

In Figure \ref{fig:supp_instruction}, we present an example of LLM summarization instructions on the YouCook2 training set. We assign the role of sentence summarizer to the LLM, imposing constraints such as length limits and prohibitions on additional explanations to ensure that the summaries adhere to a consistent format. We then provide the LLM with summarization commands and sentences as input, giving clear instructions for summarization. Through this process, we obtain semantically well-condensed summaries that form the basis of the compact memory in our proposed method.

\section{Qualitative Example}

\begin{figure*}[t]
    \centering
    \includegraphics[width=0.9\linewidth]{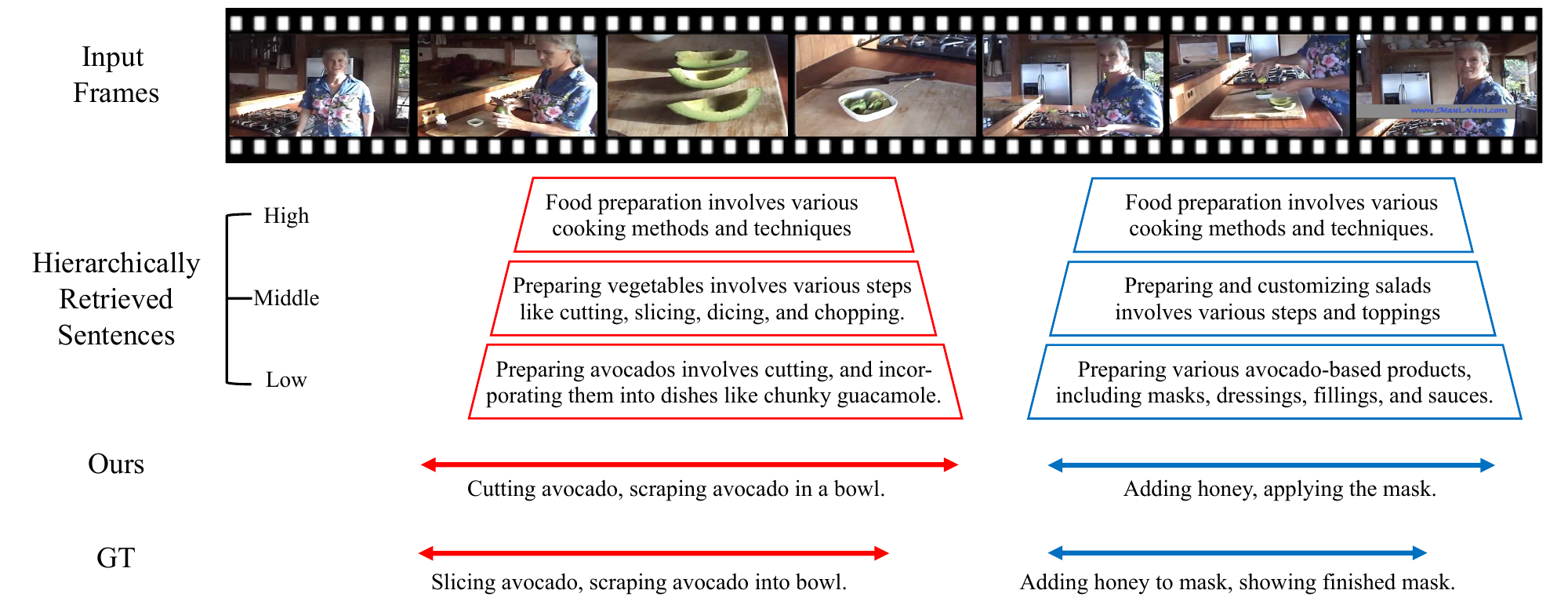}
    \caption{\textbf{Example of predictions on the ViTT test set using our approach.} The hierarchical retrieved sentences shown are examples of retrieval results with the highest semantic similarity at each hierarchical level, corresponding to specific segments of the input frames. Each retrieved sentence is converted into features and utilized in our model's predictions for the segments. Matching colors indicate the association between retrieved knowledge and prediction.}
    \label{fig:supp_qualitative}
\end{figure*}

In Figure~\ref{fig:supp_qualitative}, we show additional qualitative examples of our approach. We present an example of predictions made by our approach, highlighting the effective use of memory retrieval that draws on meaningful and relevant sentences from different memory levels, accomplished through segment-level video-text retrieval for the given video. As illustrated in the figure, high-level sentences cover broad methods or ingredients, middle-level sentences provide more detailed actions and contexts, and low-level sentences offer precise descriptions of specific subjects and actions in the scene. As a result, our method produces event boundaries and captions that are both relatively accurate, encompassing abstract concepts as well as detailed descriptions.

\end{document}